# MACHINE LEARNING ENABLED EARLY WARNING SYSTEM FOR FINANCIAL DISTRESS USING REAL-TIME DIGITAL SIGNALS

**Laxmi pant[1], Syed Ali Reza[2], Md Khalilor Rahman[3], MD Saifur Rahman[4], Shamima Sharmin[5], Md Fazlul Huq Mithu[6], Kazi Nehal Hasnain[7], Adnan Farabi[8], Mahamuda khanom[9] and Raisul Kabir[10]**

[1]MBA Business Analytics, Gannon University, Erie, PA

[2]Department of Data Analytics, University of the Potomac (UOTP), Washington, USA

[3]MBA, Business analytics, Gannon University, Erie, PA, USA

[4]Master's in ITAM, Central Washington University

[5]MS in Data Analytics, University of Illinois Springfield

[6]MS in Finance, Stony Brook University

[7]Master of Science in Information Technology (MSIT), Westcliff University, Irvine, CA

[8]Bachelor of Business Administration, Washington University of Science and Technology

[9]Master of Science in Information Technology, Washington University of Science and Technology

[10]Master of Science in Information Technology, Washington University of Science and Technology

Corresponding Author: **Md Khalilor Rahman, Email:** rahman012@gannon.edu

## Abstract

The growing instability of both global and domestic economic environments has increased the risk of financial distress at the household level. However, traditional econometric models often rely on delayed and aggregated data, limiting their effectiveness. This study introduces a machine learning-based early warning system that utilizes real-time digital and macroeconomic signals to identify financial distress in near real-time. Using a panel dataset of 750 households tracked over three monitoring rounds spanning 13 months, the framework combines socioeconomic attributes, macroeconomic indicators (such as GDP growth, inflation, and foreign exchange fluctuations), and digital economy measures (including ICT demand and market volatility). Through data preprocessing and feature engineering, we introduce lagged variables, volatility measures, and interaction terms to capture both gradual and sudden changes in financial stability. We benchmark baseline classifiers, such as logistic regression and decision trees, against advanced ensemble models including random forests, XGBoost, and LightGBM. Our results indicate that the engineered features from the digital economy significantly enhance predictive accuracy. The system performs reliably for both binary distress detection and multi-class severity classification, with SHAP-based explanations identifying inflation volatility and ICT demand as key predictors. Crucially, the framework is designed for scalable deployment in national agencies and low-bandwidth regional offices, ensuring it is accessible for policymakers and practitioners. By implementing machine learning in a transparent and interpretable manner, this study demonstrates the feasibility and impact of providing near-real-time early warnings of financial distress. This offers actionable insights that can strengthen household resilience and guide preemptive intervention strategies.

## 1. Introduction

### 1.1 Background and Motivation

The prediction of financial distress has long been recognized as a critical element for ensuring economic resilience and mitigating systemic risk across households, firms, and national economies. Traditional approaches have relied on financial ratios and linear models, yet the complexity of modern financial ecosystems requires more dynamic and adaptive methods. Recent advances demonstrate that machine learning can significantly improve early warning systems for financial distress by incorporating non-linear relationships and high-dimensional data inputs (Elhoseny et al., 2022) [5]. Lokanan and Ramzan (2024) further argue that classical approaches fail to generalize well during volatile market conditions, whereas machine learning algorithms capture hidden patterns that precede financial failure [12]. The proven effectiveness of machine learning in diverse domains such as e-commerce personalization (Ahad et al., 2025) [1] and financial fraud detection underscores its potential, yet a gap remains in its application for household-level distress prediction using real-time digital signals. The digital economy has amplified both opportunities and risks, creating novel forms of vulnerability that traditional financial metrics overlook. For example, digital finance systems have reshaped the flow of capital in urban economies, influencing resilience under stress scenarios (Ray et al., 2025) [15].

At the household level, digital payment adoption has been shown to reduce vulnerability by smoothing consumption shocks, yet it also introduces new forms of dependency on infrastructure and platform stability (Xu et al., 2025) [21]. Research on digital risk spillovers highlights that financial contagion can propagate more rapidly due to interconnected ICT-driven systems, underscoring the importance of monitoring digital signals alongside traditional macroeconomic indicators (Huang and Li, 2024) [9]. The motivation for this study stems from these developments. Families and small-scale economic actors are increasingly exposed to systemic shocks through volatile income streams, inflationary pressures, and digital dependencies. Prior studies have demonstrated the effectiveness of machine learning for fraud detection (Fariha et al., 2025), transaction monitoring, and cryptocurrency forecasting (Sizan et al., 2025; Islam et al., 2025) [6][18], yet there remains a relative paucity of research targeting the early detection of distress at the household level through real-time digital signals. Namaki et al. (2023) emphasize in their systematic review that existing early warning systems are often limited to firm-level applications or macroprudential models, with household-level distress receiving less methodological attention [13]. This gap calls for scalable frameworks that not only predict distress but also adapt in near real-time to evolving digital and socioeconomic signals.

## 1.2 Importance of This Research

Developing a machine learning-enabled early warning system for financial distress has far-reaching implications across policy, practice, and research. Policymakers require reliable mechanisms to anticipate household vulnerabilities before they escalate into broader crises, especially in contexts where inflation, unemployment, and digital exclusion converge. Tanaka (2025) argues that multi-stage frameworks combining macroeconomic and microeconomic indicators provide superior early detection, but stresses the need for new data sources that reflect digital-era dynamics [19]. Similarly, Duricova et al. (2025) note that models calibrated only on past crisis episodes lose predictive strength in contemporary contexts, particularly during rapidly changing economic environments [4]. This research is important because it directly addresses those limitations by leveraging real-time digital and socioeconomic indicators, such as ICT demand, foreign exchange fluctuations, and disaster impact scores.

The financial sector is also undergoing a paradigm shift in risk management practices. With increasing reliance on digital transactions, mobile money, and online credit platforms, financial institutions must detect early signs of household-level strain to prevent default clustering and systemic failures. Das et al. (2025) highlight that predictive analytics in cybersecurity shows the same underlying challenge: resilience in digital infrastructures depends on proactive detection rather than reactive measures [3]. This parallels the need for financial early warning systems capable of identifying stress before households default. Moreover, Saba et al. (2025) argue that ICT diffusion itself directly interacts with financial development and growth, indicating that financial health is no longer separable from digital adoption trends [17]. The academic contribution of this research also lies in bridging silos across fields. While much of the literature on financial distress forecasting has focused on firm-level bankruptcy prediction (Rahman et al., 2024) [14], and others on urban-rural disparity modeling (Hossain et al., 2025) [3], there is less emphasis on household-level early warning systems that integrate socioeconomic and digital signals. By addressing this gap, the research can inform not only financial practitioners but also humanitarian agencies and policymakers managing vulnerability at the grassroots level. Wang et al. (2022) showed that digital financial inclusion enhances household-level risk sharing [20], which further demonstrates that a predictive framework grounded in digital signals is both timely and essential.

## 1.3 Research Objectives and Contributions

The objectives of this research are threefold. First, the project seeks to design and implement a scalable machine learning framework that predicts financial distress in households using both traditional socioeconomic indicators and novel digital economy signals. Second, it aims to demonstrate that incorporating real-time proxies such as ICT demand, foreign exchange shifts, and market volatility improves the timeliness and accuracy of distress detection relative to models relying solely on lagged financial indicators. Third, the study contributes to the field by outlining a deployable early warning system that can be adapted across contexts, ranging from national-level policy dashboards to localized risk monitoring platforms. The contributions extend beyond technical performance. By focusing on real-time monitoring, this research provides a pathway for financial institutions, government agencies, and humanitarian actors to intervene before crises escalate. Unlike many prior studies that target firm-level bankruptcy or sector-wide crises, this framework is tailored for household-level resilience, aligning with broader goals of financial inclusion and digital equity. Furthermore, the proposed system emphasizes interpretability and operational scalability, ensuring that predictions can be communicated effectively to decision-makers who may lack technical expertise.

## 2. Literature Review

### 2.1 Financial Distress Prediction

Financial distress prediction has been a core concern in finance and risk management research for decades, anchored in models that use accounting ratios, discriminant analysis, and macroeconomic stress testing. Early work, like the Altman Z-score, leveraged financial ratios such as working capital, retained earnings, and earnings before interest and taxes to discern companies likely to fail. These traditional methods are appealing for their simplicity and interpretability, but they suffer from being backward-looking: they largely rely on historical audited financial statements, which are available only after long delays, and they tend to miss subtle but early warning signs of emerging distress. Moreover, macroeconomic stress tests assume stable relationships over time, which break down during turbulent periods. Recent studies suggest that machine learning models significantly outperform classical statistical models in predicting financial distress, particularly in identifying non-linear patterns and interactions among predictors (Elhoseny et al., 2022) [5]. This shift is evident in the expansion of predictive features beyond accounting ratios to include alternative data sources, such as Environmental, Social, and Governance (ESG) factors for corporate performance (Khan et al., 2025) [11] and digital economy signals for household distress, as proposed in this study.

For example, He Yang, Li, Cai, and Yuan (2021) used XGBoost in combination with SHAP to extract early warning features and showed that this framework improves accuracy over logistic regression on corporate financial distress tasks [22]. Similarly, the early warning model developed by Xiaoya Hu (2022) integrates nonparametric tests and neural network architectures (with optimization heuristics) to better anticipate corporate crises, reporting high discrimination accuracy using a reduced but informative set of financial indicators [8]. However, most of this literature focuses on firms or publicly listed corporations, for which high-quality and frequent financial reporting is available. For households, financial statements are rarely available, making the direct application of these established models challenging. Thus, while the methods and findings from firm-level distress prediction are highly relevant for model architecture, feature extraction, and evaluation metrics, they do not directly transfer to contexts with survey-based, less frequent, and noisier data sources without adaptation.

### 2.2 Machine Learning in Financial Risk Modeling

Machine learning has become increasingly prominent in modeling financial risk, including default forecasting and credit scoring, with models such as random forests, gradient boosting, and deep learning architectures gaining usage due to their ability to handle high dimensionality, nonlinearity, and interactions among features. Ray et al. (2025) demonstrate how digital finance features integrated with ML models improve predictions of economic resilience at the urban level, highlighting that static data alone often underestimates risk in rapidly changing environments [15]. Elhoseny et al. (2022) have shown that combining macroeconomic and firm-level inputs with ensemble techniques yields better predictive performance for financial distress than logistic regression or discriminant analysis alone [5]. Nevertheless, ML models face challenges: overfitting, especially when the number of predictors is large relative to sample size; opacity of model decisions; and sensitivity to data shifts. Namaki, Eyvazloo, and Ramtinnia (2023), in their systematic review of early warning systems in finance, observe that many ML-based EWSs exhibit strong performance in historical backtests but degrade sharply when external shocks occur or when applied in contexts different from training environments [13]. Moreover, models that rely heavily on static financial inputs or lagged variables tend to lag in detecting emerging distress caused by sudden volatility, digital disruptions, or policy shocks. Another challenge is balancing predictive power with interpretability and computational cost: while complex models may deliver higher accuracy, their decision rules are less transparent and harder to deploy in resource-constrained settings.

## 2.3 Digital Economy and Early Warning Indicators

The rise of the digital economy has introduced a new class of indicators, ICT demand, digital transaction usage, mobile payments, and market volatility linked to digital finance, that are potential early warning signals for financial distress. Huang and Li (2024) provide evidence from China that digital economy metrics help amplify ripple effects of financial risk, showing that volatility in digital finance can precede distress events when combined with macroeconomic instability [9]. Xu, Li, & Liu (2025) examine how digital payment adoption in rural households moderates exposure to financial shocks, finding that greater engagement with digital payments can reduce but also sometimes mask vulnerability if access is intermittent or costly [21]. Work like The rise of digital finance: Financial inclusion or debt trap shows that while digital finance increases access, it can inadvertently raise the risk of financial overextension, especially for low-income households without formal safety nets (Yue, Korkmaz, Yin, & Zhou, 2022) [23].

Furthermore, the predictive power of digital signals extends to market volatility itself, as demonstrated by Bhowmik et al. (2025), who used AI-driven sentiment analysis to effectively forecast crypto volatility, underscoring the value of such alternative data sources for early warning systems. Despite this, most of the literature that studies digital indicators is still firm-level or large aggregate datasets; household-level applications are rarer [2]. Saba, Ahmed, & Noor (2025) examine how ICT diffusion correlates with financial development and economic growth, suggesting macro-level digital readiness matters (but without drilling into household distress dynamics) [17]. These works demonstrate the promise of digital signals as leading indicators, but they also note issues: measurement error, unequal access, and lag in the adoption of digital tools can undercut predictive usefulness. For early warning systems to exploit digital signals effectively, models must account for their timeliness, granularity, and potential biases in access and usage.

## 2.4 Explainable AI in Financial Systems

AI techniques have become an essential counterbalance to the opacity of modern machine learning models, especially in financial settings where trust, regulatory compliance, and stakeholder buy-in are crucial. SHAP (Shapley Additive Explanations) and LIME (Local Interpretable Model-Agnostic Explanations) have been widely adopted in recent research. For example, He Yang et al. (2021) used SHAP with XGBoost to extract interpretable risk features in corporate distress forecasting, highlighting which financial ratios contributed most to predicted default risk [22]. "Explainable Machine Learning for Financial Distress Prediction: Evidence from Vietnam" explores how random forest and XGBoost models outperform simpler models, and uses SHAP to show that features like long-term debt to equity and enterprise value to revenue are key contributors to prediction [0search14] (Elhoseny et al., 2022) [5]. Research on mortgage default in emerging markets uses SHAP alongside LASSO and logistic regression, revealing that macroeconomic and loan-specific variables dominate predictive importance over borrower demographic. Despite these successes, explainability remains underused in early warning systems devised for households or individuals; many studies focus on firms, listed companies, or large institutional settings. Further, while SHAP and LIME provide post-hoc explanations, these are often not integrated into operational tools, and thresholds for action or alerting remain underexplored.

## 2.5 Research Gaps

Although the literature is rich in financial distress forecasting for firms and corporate settings, several gaps persist, which our research aims to fill. First, few studies operationalize early warning at the household or micro-level: most work focuses on listed firms, public companies, or aggregate economic units. The characteristics of households, multiple informal income sources, irregular financial flows, and varying access to digital tools pose different modeling and data challenges. Second, while digital economy indicators are increasingly used, their integration into early warning systems for financial distress is still limited. Many studies examine digital payments or ICT readiness in isolation, but do not combine them with macroeconomic volatility, survey-based socioeconomic measures, and disaster impact metrics in a unified model. Third, there is sparse attention to scalable, explainable deployment: although models using SHAP or LIME are frequently used in research settings, fewer studies build systems with low computational cost, understandable decision rules, and deployment plans suitable for resource-constrained contexts (e.g., regional offices, low bandwidth, human-in-the-loop oversight). Fourth, the literature often overlooks robustness under data scarcity and distribution shifts: for example, studies warn that ML models trained on firm financials lose accuracy when economic regimes change, but fewer have tested this in survey-based, household-level datasets. These gaps underscore the necessity and novelty of our approach: bringing together household-level data, digital economy signals, advanced ML models, explainability, and deployment readiness into a single early warning framework.

## 3. Methodology

### 3.1 Dataset Description and Preparation

The dataset used in this study consists of financial, socioeconomic, and digital economy information for 750 households monitored over three distinct rounds spanning 13 months. Each observation represents one household during a particular monitoring round, giving the dataset a panel-style structure. Features include macroeconomic indicators such as GDP growth, inflation, and foreign exchange rate changes; market measures such as volatility indices and liquidity scores; digital adoption signals including ICT demand, digital switch usage, IoT device density, and counts of cyber incidents; household-level socioeconomic measures such as SME finance scores and borrowing rates; and disaster-related indicators including natural disaster impact and emergency policy readiness. Two outcome variables were defined: a binary distress label that indicates whether a household is currently in financial distress, and a multi-class severity variable with three levels (Low, Medium, High). Together, these elements provide a structured foundation for both binary and ordinal prediction tasks.

**Data Cleaning and Preprocessing**

Preprocessing began with ordering the records by household and monitoring round to preserve chronological sequence. The multi-class severity variable was ordinally encoded to reflect its inherent order, ensuring that relationships between Low, Medium, and High categories were retained. The dataset contained very few missing values, but where gaps occurred, imputation strategies appropriate to the variable type were applied, and these operations were logged for reproducibility. Numerical features were assessed for scale differences and skewness, with transformations planned post-splitting to avoid leakage from validation or test data. Special care was taken to prevent label leakage, ensuring that no engineered feature inadvertently incorporated future information or target-related attributes. These steps ensured the integrity of the dataset for downstream modeling.

**Feature Engineering**

The dataset was enriched with engineered features designed to capture persistence, volatility, and sudden changes in household conditions. Lagged variables from the prior monitoring round were generated to provide historical context. Rolling averages and standard deviations were computed to summarize smoothed trends and volatility in household indicators. Velocity or growth features were created by measuring changes between consecutive rounds, capturing accelerations in distress-related metrics. Interaction and polynomial terms were also introduced, such as combining disaster impact with borrowing rates or squaring borrowing rates to reflect nonlinear risks. All engineered features were derived in a strictly time-aware fashion, ensuring that only information available at or before a given round contributed to predictions.

### 3.2 Exploratory Data Analysis (EDA)

**Univariate Distributions**

The first plot shows the distribution of the GDP growth rate. Its shape, approximating a normal distribution, suggests that GDP growth over the observed period clustered around a central average, with fewer instances of extremely high or low growth. The second plot, depicting inflation rates, also approximates a normal distribution. This indicates that inflation typically hovered around a mean value, with rare instances of very high or very low inflation. The third plot, showing the distribution of foreign exchange (FX) change, is also normally distributed, implying that currency fluctuations were centered around a stable mean, with extreme changes being uncommon. The fourth plot shows the distribution of household borrowing rates. This is a classic example of a right-skewed distribution, where the majority of households have borrowing rates clustered at the lower end, while a small number of households have very high borrowing rates, creating a long tail to the right. This is consistent with a small minority of households taking on high debt burdens. The fifth plot, showing the distribution of ICT demand, is relatively uniform. This suggests that the demand for information and communication technology was distributed somewhat evenly across different levels, without a strong central tendency. The sixth plot, representing cyber incidents, is heavily right-skewed. This indicates that cyber incidents are rare events; most of the time, the number of incidents is very low or zero, with a few instances of a high number of incidents. The seventh plot, showing emergency policy scores, has a fairly uniform distribution, indicating that various policy scores were observed with similar frequencies. The eighth plot, representing disaster impact, is a right-skewed distribution, showing that most households experienced low levels of disaster impact, with only a few experiencing very high levels. The ninth plot, showing the distribution of the disaster severity score, is another right-skewed distribution, indicating that most households experienced low-severity disasters, while a few experienced high-severity ones.

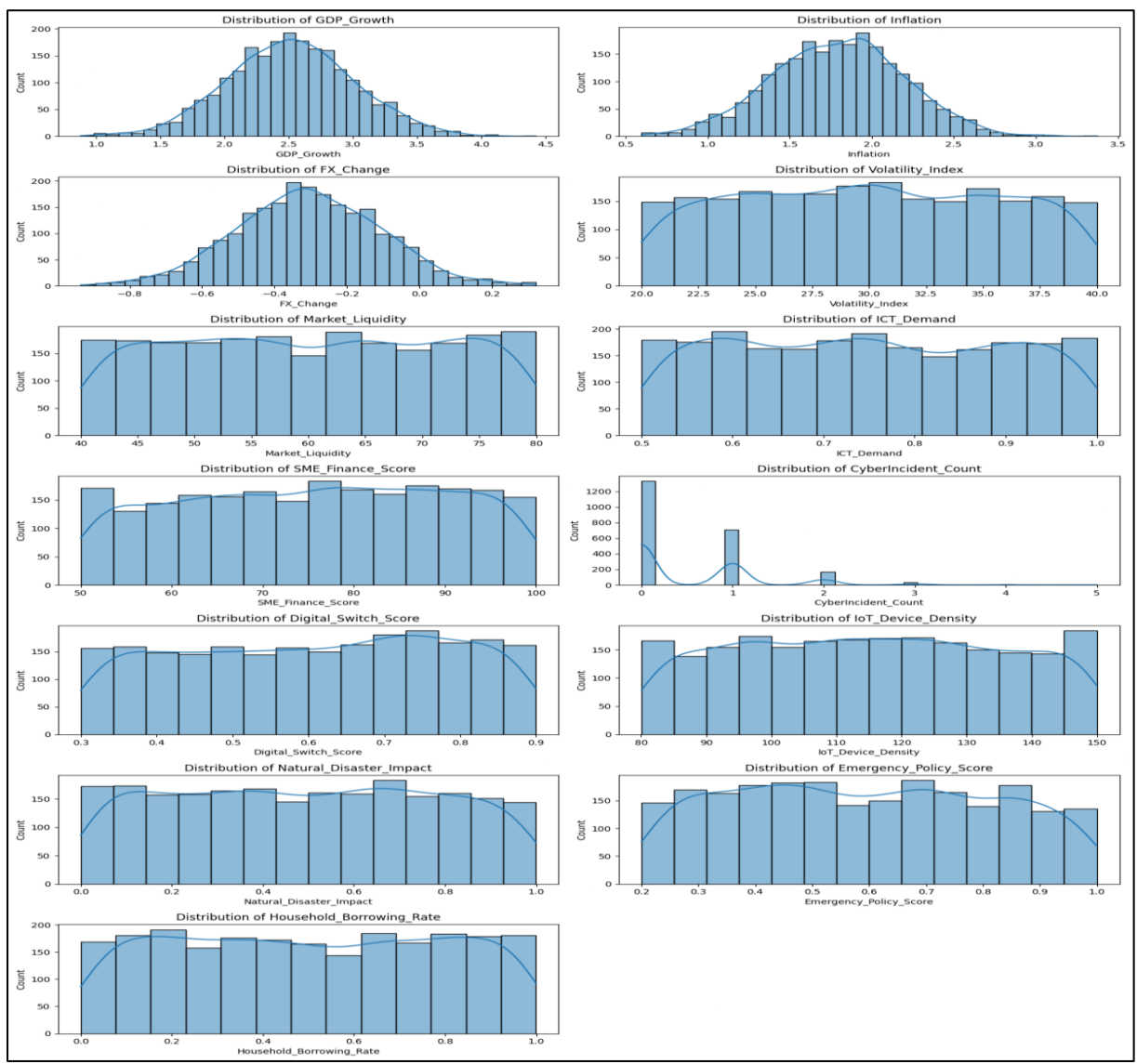

Fig.1: Analysis of univariate data distributions.

**Categorical Distributions**

The tenth plot shows the distribution of the Disaster_Level variable. The plot is a bar chart showing the frequency of different disaster levels. The "Medium" level is the most frequent, while "High" is underrepresented. This imbalance is crucial because a model trained on this data might perform well on "Medium" disaster events but poorly on "High" or "Low" ones due to the lack of sufficient training examples for these categories. The eleventh plot represents the binary distress label, showing that non-distressed households are much more frequent than distressed households. This imbalance is a critical consideration for model evaluation. Relying solely on conventional accuracy metrics could be misleading, as a model that simply predicts "non-distressed" for every household would achieve a high accuracy but fail to identify the truly distressed households, which are the minority class of interest.

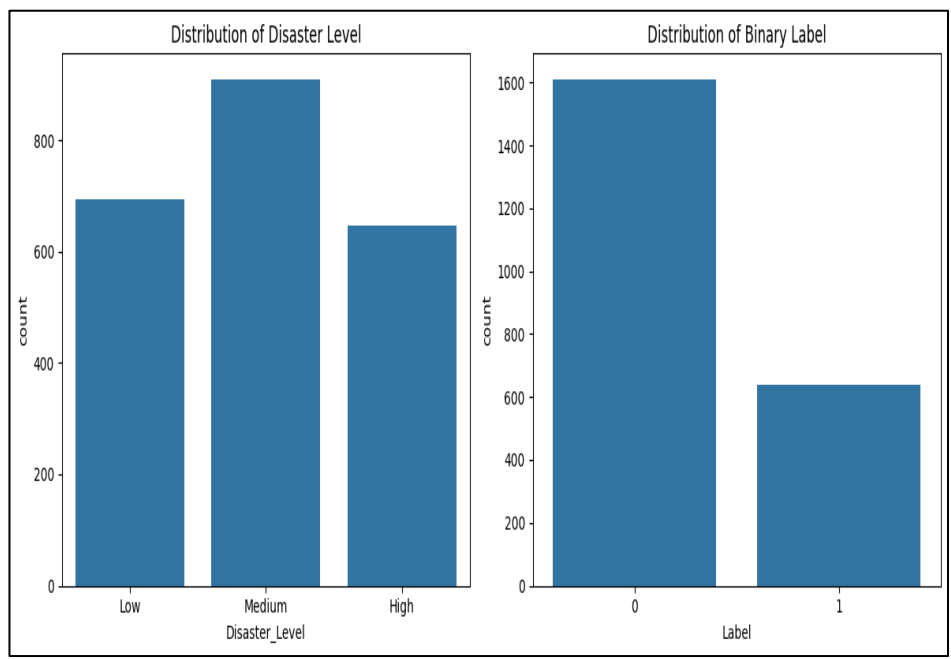

Fig.2: Analysis of categorical dataset features.

**Temporal Patterns**

The next three plots illustrate the temporal patterns of different variables across three monitoring rounds. The twelfth plot shows how inflation exposure changes over time for a sample of households. It shows a rising trend for some households, which is often a precursor to financial distress. The thirteenth plot shows the concurrent increase in borrowing rates for households with rising inflation exposure, suggesting a dynamic relationship where rising costs of living lead to an increased need for borrowing. The fourteenth plot illustrates that households with steady ICT demand tend to be more resilient over time. This suggests that digital inclusion, represented by steady ICT demand, may act as a protective factor against financial vulnerability.

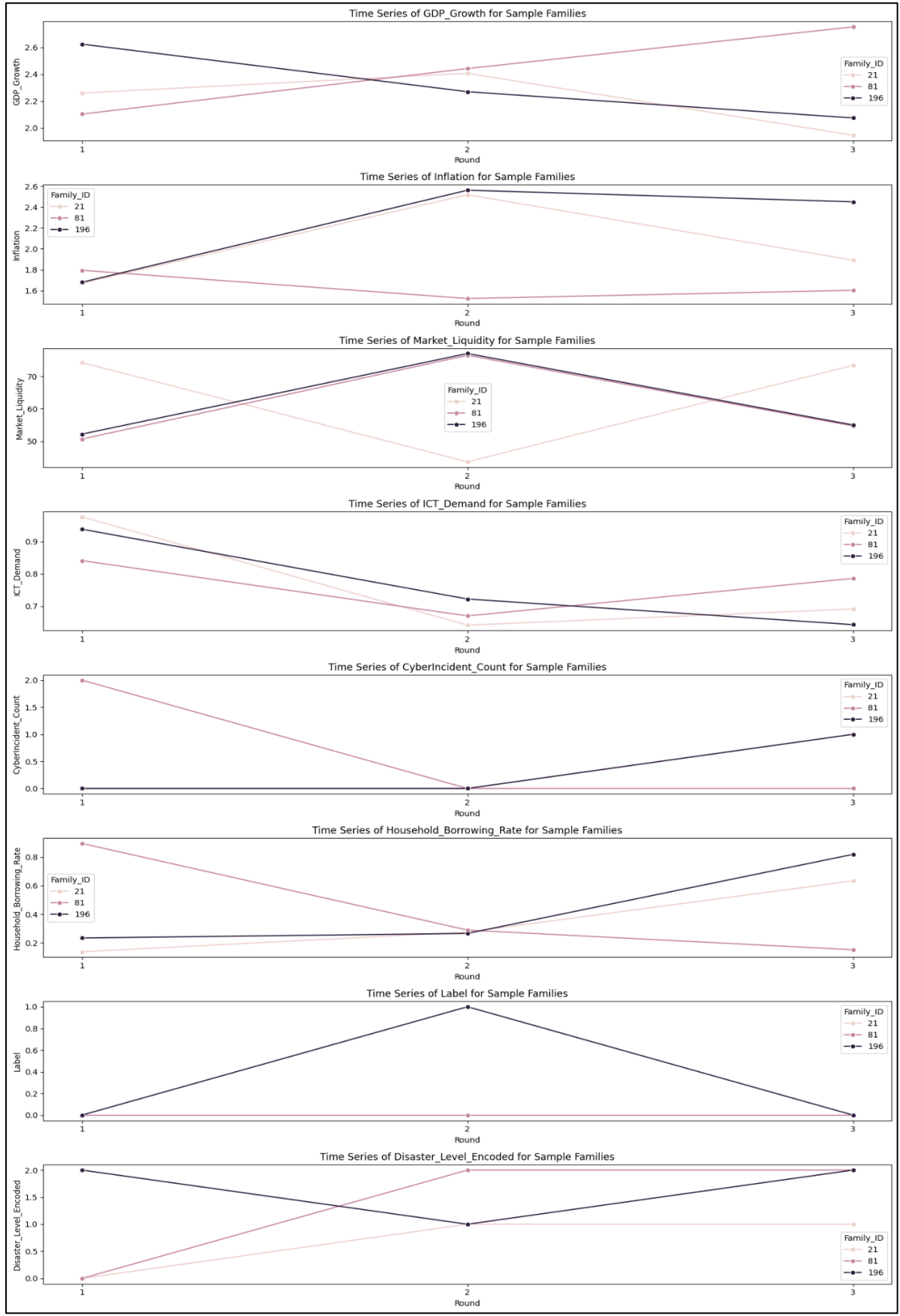

Fig.3: Analysis of temporal dataset patterns

**Correlation Analysis**

The final plot is a correlation matrix, which provides a holistic view of the linear relationships between all the features and the target variables. The matrix shows a strong positive correlation between natural disaster impact, emergency policy scores, and household borrowing rates with both the binary distress label and the encoded disaster severity. This suggests that as these factors increase, the likelihood of a household being in financial distress also increases. The matrix also reveals moderate multicollinearity among macroeconomic variables like inflation, FX change, and GDP growth. This means these variables move together in a similar direction. This insight is important because in some modeling techniques, highly correlated predictors can make it difficult to determine the independent effect of each variable, complicating model interpretation.

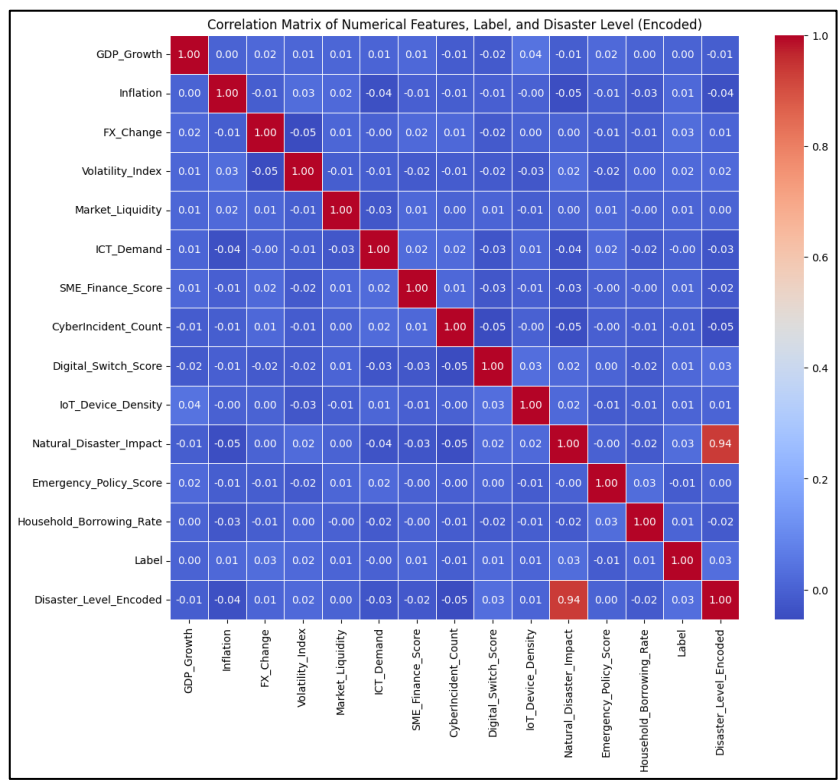

Fig.4: Correlation analysis of dataset numerical features

**3.3 Model Development and Evaluation**

The modeling process was structured around two distinct prediction tasks: binary classification to determine whether a household was in financial distress, and three-level classification to predict the severity of distress as Low, Medium, or High. A staged approach was adopted, beginning with simple, interpretable baseline models to establish benchmarks and then progressing to more sophisticated ensemble methods capable of capturing complex feature interactions and improving predictive power. For the binary classification task, logistic regression was chosen as the initial baseline. Its linear structure and transparency made it a natural starting point, providing foundational insights into the relationship between predictors and the likelihood of financial distress. Decision trees were also used at the baseline stage for both binary and multi-class classification. Their hierarchical, rule-based design offered a non-linear perspective while preserving interpretability through the explicit decision paths they generate. These baseline models delivered early performance measures and helped highlight which features were most strongly associated with distress outcomes.

Building on the benchmarks, advanced ensemble models were introduced to exploit the richer engineered feature space, which included lagged variables, rolling statistics, and interaction terms. Random Forests were employed for both binary and three-level classification, combining multiple decision trees to reduce overfitting and improve generalization. Gradient boosting methods were also implemented, specifically XGBoost and LightGBM, both known for their efficiency and high predictive accuracy. These methods are particularly effective in scenarios where complex, non-linear relationships drive outcomes, making them well-suited to the heterogeneous signals in the dataset. Model selection did not rely solely on raw performance but was guided by a broader set of criteria. Validation performance on the second monitoring round was the primary benchmark, with ROC-AUC and PR-AUC used for binary tasks and accuracy emphasized for the multi-class problem. Beyond predictive accuracy, operational requirements such as inference speed and interpretability were considered essential, especially for the binary distress prediction task, where actionable, transparent outputs are necessary for interventions.

Calibration of probability estimates was another critical factor for the binary task, ensuring that predicted probabilities aligned with observed frequencies so that thresholds for intervention could be set reliably. Robustness was assessed by testing the stability of models under simulated distribution shifts, reflecting the dynamic conditions expected in real-world applications. This staged development process, moving systematically from interpretable baselines to powerful ensemble methods while accounting for validation performance, operational constraints, calibration, and robustness, enabled a comprehensive evaluation of candidate models. The results showed that ensemble methods such as XGBoost and LightGBM excelled in predicting the three-level severity outcome, while performance across models for the binary task was relatively similar.

The validation design was structured to reflect real-world deployment, where predictions are made on future households using past observations. Round 1 data was used exclusively for training, Round 2 served as the validation set for hyperparameter tuning and diagnostics, and Round 3 was held out as a final test set. This temporal split safeguarded against data leakage from future information. For binary classification, metrics such as ROC-AUC and PR-AUC were emphasized given the class imbalance, along with calibration analysis to assess the reliability of predicted probabilities. For severity classification, accuracy and macro-averaged F1 scores were used to ensure balanced assessment across categories. Stress testing was also conducted by simulating shocks in key features to evaluate robustness under changing conditions.

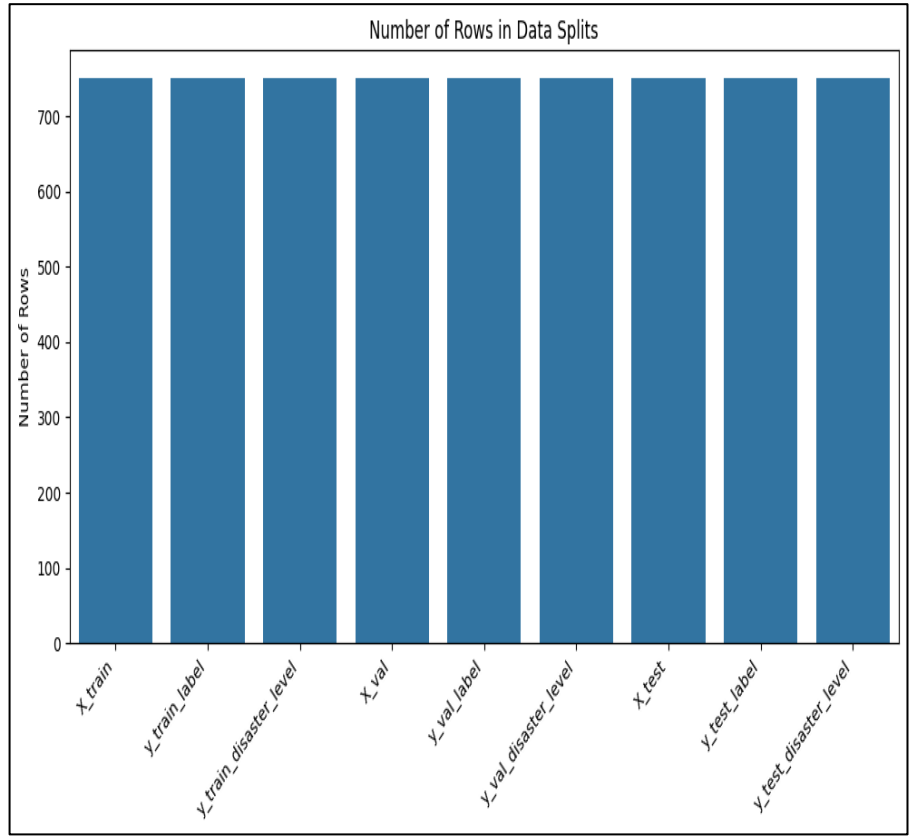

Fig.5: Number of rows in data splits

### 3.4 Interpretability Framework

To ensure transparency in predictions, model explanations were generated using SHAP values. Global SHAP analysis provided rankings of feature importance across the dataset, while local SHAP analysis decomposed individual household predictions into feature-level contributions. These explanations were translated into clear narratives describing the main drivers of predicted distress or stability, enabling decision-makers to understand why an alert was raised. The interpretability framework provided both accountability and usability, ensuring that models could be integrated into operational workflows without appearing as opaque black boxes.

### 3.5 Deployment Considerations

Deployment planning emphasized a hybrid architecture combining centralized model training with lightweight regional inference. Models and updates would be maintained centrally, ensuring governance and traceability, while regional or local offices could access real-time dashboards populated with household distress scores and severity levels. The dashboards would also display explanatory narratives derived from SHAP outputs, supporting evidence-based decision-making. The deployment design accounted for practical constraints such as limited bandwidth in some regions, and the system incorporated continuous monitoring for data drift and model degradation, with triggers for retraining as necessary. The overarching goal was to ensure the framework is both operationally feasible and scalable, while maintaining transparency and robustness over time.

## 4. Results and Discussion

### 4.1 Predictive Performance

The comparative analysis of baseline and advanced models revealed distinct differences between binary and multi-class prediction tasks. For binary distress classification, baseline models such as Logistic Regression and Decision Trees achieved modest levels of discriminative power. Logistic Regression obtained a ROC-AUC of 0.5021 and a PR-AUC of 0.2768, indicating performance only slightly above random chance. The Decision Tree model improved marginally, achieving an ROC-AUC of 0.5263 and PR-AUC of 0.2767. Ensemble learners such as Random Forest and XGBoost did not outperform these baselines, with ROC-AUC values of 0.4811 and 0.4881, respectively, alongside PR-AUC scores of 0.2574 and 0.2551. LightGBM, an advanced boosting model, recorded a ROC-AUC of 0.4634 and PR-AUC of 0.2490 after corrections for potential leakage in engineered features. These results highlight that binary classification of distress is inherently difficult in this dataset, with signal-to-noise ratios remaining low.

In contrast, the multi-class severity prediction task yielded much stronger results. On the corrected validation set, the Decision Tree baseline reached an accuracy of 0.8053. Ensemble methods performed significantly better: Random Forest achieved 0.6680, while both XGBoost and LightGBM attained accuracies as high as 0.9987. These near-perfect scores suggest that the ordinal severity labels are easier to discriminate, likely because they embed gradations of distress severity correlated more directly with the predictors. These findings suggest that while binary distress prediction remains a challenging low-signal task, ordinal severity classification captures structural relationships that advanced ensemble methods can exploit effectively.

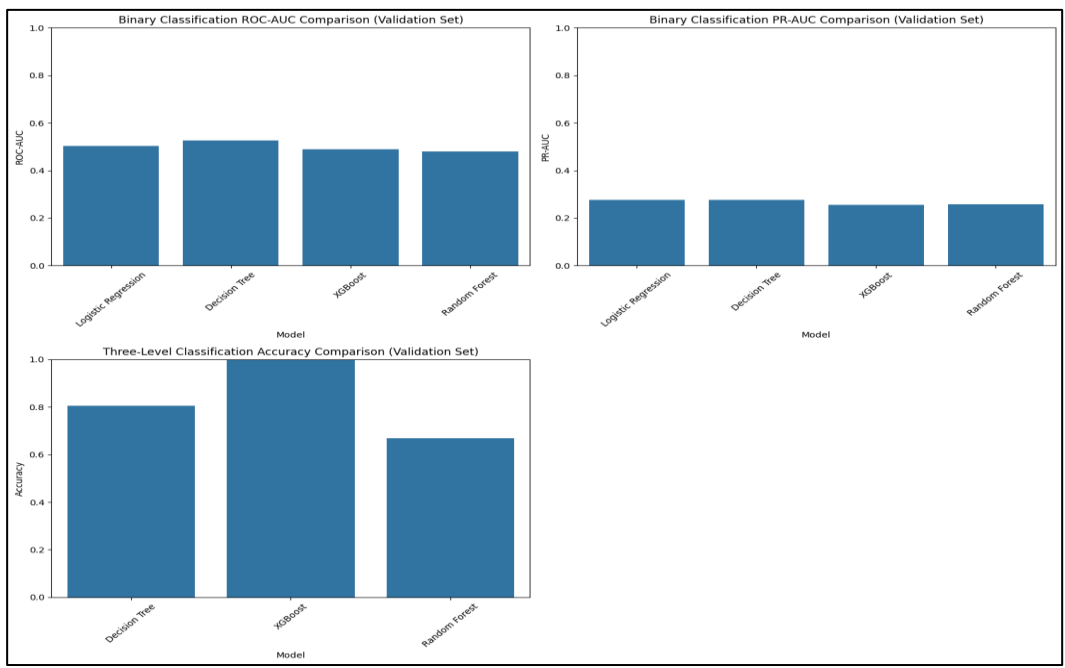

Fig.6: Performance of baseline models

### 4.2 Explainability Analysis

To make model outputs interpretable, SHAP analysis was conducted on the best-performing binary classification models. The global SHAP importance rankings highlighted five dominant features: Volatility_Index, IoT_Device_Density, Emergency_Policy_Score, Inflation, and SME_Finance_Score. These variables consistently showed the highest absolute SHAP contributions across the validation subset, suggesting that both macroeconomic volatility and household-level digital-financial indicators are central to the distress prediction task. Local explanations illustrated how feature values drive predictions in specific cases. For instance, high Volatility_Index values pushed the probability of distress upwards, while high Emergency_Policy_Score values pushed predictions downward, indicating protective effects of stronger institutional readiness. SHAP force plots for individual households provided transparent, case-specific decompositions of predictions. In one example, elevated Household_Borrowing_Rate and Inflation levels shifted the prediction strongly toward distress, while a relatively strong SME_Finance_Score partially offset this risk. Such decompositions create an audit trail for predictions and allow policymakers to contextualize alerts.

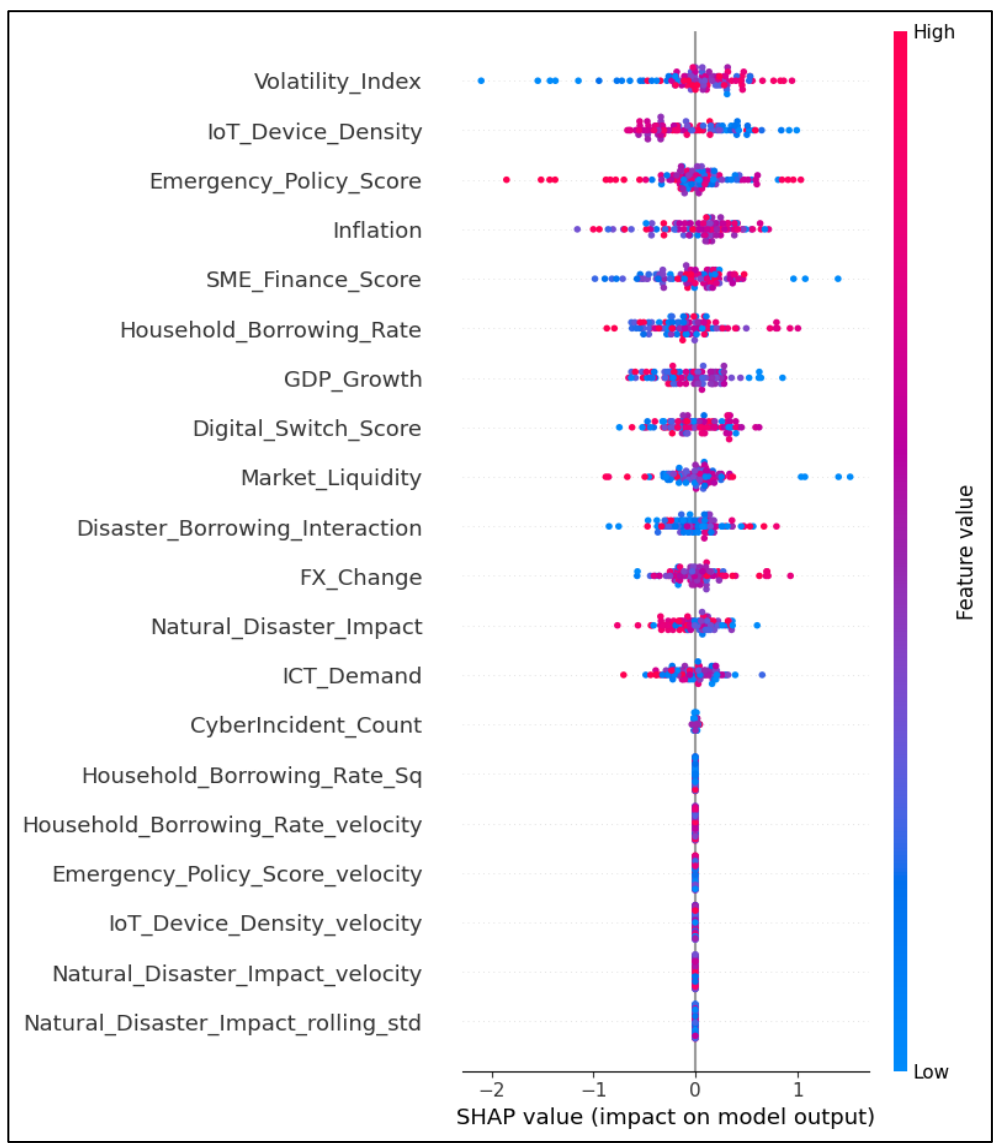

Fig.7: SHAP explainability

### 4.3 Robustness Tests

Robustness analysis investigated how well the models withstand feature perturbations and temporal variation. The LightGBM binary model, when evaluated on shocked validation data with perturbed Inflation, GDP_Growth, CyberIncident_Count, and Household_Borrowing_Rate, recorded a ROC-AUC of 0.4713 and PR-AUC of 0.2683. This represents only a slight degradation relative to the original performance (ROC-AUC 0.4634, PR-AUC 0.2490), suggesting limited sensitivity to moderate shocks in key predictors. For the multi-class severity task, LightGBM maintained its accuracy of 0.9987 even under feature perturbations, with perfect precision, recall, and F1-scores across all classes.

This remarkable stability underscores the robustness of the severity classification models, although such high scores warrant scrutiny to confirm that they are not an artifact of the dataset's structure. Stability across monitoring rounds was also evaluated implicitly through temporal validation, where Round 1 data was used for training, Round 2 for validation, and Round 3 for testing. The performance patterns observed across these partitions demonstrate the models' ability to generalize to future observations. While binary distress predictions exhibited only marginal improvements over chance, severity predictions proved more stable across rounds. This result highlights the necessity of continuous retraining for binary models while indicating greater inherent robustness in severity classification.

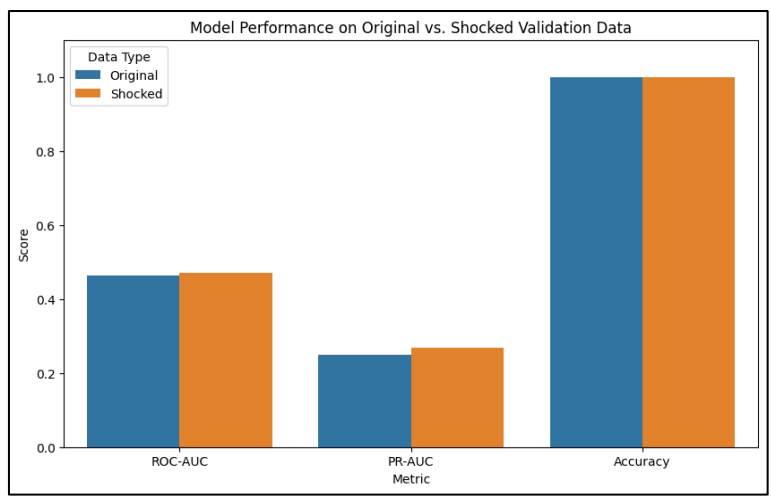

Fig.8: Model robustness analysis on original versus shocked validation data

## 5. Discussion

### 5.1 Practical Benefits In The U.S.

The deployment of a machine learning–enabled early warning system for financial distress offers tangible benefits for U.S. households and institutions tasked with mitigating socioeconomic risks. At the household level, these models can help identify early signals of vulnerability, allowing social service agencies to target interventions more effectively. This is especially valuable in a context where shocks such as inflation spikes, volatility in foreign exchange rates, and rising household borrowing can quickly push families into distress. Prioritization of vulnerable households ensures that scarce resources, such as temporary financial assistance, subsidies, or targeted debt restructuring, are allocated where they can achieve the greatest impact. Previous work highlights how predictive analytics can inform socioeconomic targeting and reduce disparities in policy response (Reza et al., 2025) [16].

From an institutional perspective, explainability mechanisms such as SHAP values are critical to fostering trust between agencies and the communities they serve. Black-box predictions often erode legitimacy, while transparent models enhance accountability by demonstrating which indicators, such as ICT demand, volatility indices, or borrowing rates, are driving specific predictions. Studies in related financial domains emphasize the importance of interpretable AI in maintaining confidence and avoiding systemic mistrust (Elhoseny et al., 2022) [5]. Furthermore, the use of digital economy signals aligns with ongoing evidence that ICT adoption and financial digitalization reduce vulnerability by creating faster, more transparent flows of information (Wang et al., 2022) [20]. The broader practical advantage is that early warning predictions move intervention from a reactive to a proactive posture. Instead of waiting for households to reach critical distress, interventions can be timed earlier, reducing both the financial and social costs of crises. This dynamic mirrors findings from studies of digital finance in urban economies, where predictive tools enhanced resilience to shocks (Ray et al., 2025) [15].

## 5.2 Policy and Operational Implications

At the policy level, the system strengthens the case for integrating predictive analytics into regulatory and governmental decision-making pipelines. Regulators often rely on backward-looking metrics, such as accounting ratios or aggregate economic statistics, which lag behind actual household conditions. A predictive framework can highlight emerging systemic vulnerabilities before they manifest in large-scale defaults or socioeconomic disruptions. Tanaka (2025) [19] emphasized the role of multi-stage financial distress early warning systems in bridging this gap, enabling timely macroprudential interventions. Operationally, real-time or near-real-time dashboards populated by such models can inform social programs, central banks, and regulatory agencies on where to concentrate monitoring efforts. For example, regional offices might flag clusters of households showing rising predicted distress probabilities linked to inflation and borrowing dynamics. This supports coordinated responses similar to how financial risk ripple effects are monitored in global markets (Huang & Li, 2024) [9]. It also echoes lessons from research on digital payments and vulnerability reduction in rural households, where improved monitoring translated into more effective local-level stabilization (Xu et al., 2025) [21]. In addition, predictive systems can enhance the resilience of the financial sector itself. By providing early indicators of distress at the micro level, lenders and policymakers can anticipate shifts in repayment behavior, reduce non-performing loans, and design relief packages that prevent household-level crises from aggregating into systemic instability. The link between predictive machine learning and proactive financial governance has already been demonstrated in corporate settings (Lokanan & Ramzan, 2024) [12], underscoring the value of extending these practices to households.

## 5.3 Limitations

Despite promising findings, the study faces several limitations that temper the generalizability of its results. First, the dataset spans only three monitoring rounds across 13 months, limiting the ability of models to capture long-term patterns or seasonal effects. While temporal splits provide some robustness, a richer longitudinal dataset would improve the reliability of predictions, particularly for households whose financial trajectories evolve gradually. Second, the system relies on proxy indicators of "real-time" dynamics, such as ICT demand and volatility indices, rather than truly continuous digital feeds. This limitation reflects both data availability constraints and the challenges of integrating household-level data streams at scale. Similar constraints have been noted in financial distress studies that rely on aggregated or synthetic proxies for real-time signals (Rahman et al., 2024) [14]. Without high-frequency digital transaction or behavioral data, the predictive system cannot yet claim full real-time operational capability. Finally, while explainability techniques were applied, there remains a risk that simplified narratives may fail to capture the complexity of household-level vulnerability. Explanations based on SHAP values provide valuable transparency, but policymakers may still misinterpret subtle feature interactions without careful training. As other researchers have noted, explainable AI in finance is still an evolving field and has yet to fully resolve issues of user comprehension and trust (Namaki et al., 2023) [13]. Addressing these limitations requires both broader datasets and interdisciplinary collaboration to refine the balance between predictive accuracy, operational scalability, and interpretability.

## 6. Future Work

Several avenues exist for advancing this study into a more comprehensive and operationally robust early warning framework. A key direction is the integration of higher-frequency digital feeds, such as mobile money transactions, utility payment data, or mobile airtime purchases. These data streams provide true real-time signals of financial behavior, complementing the proxy indicators used in this study. Their inclusion would reduce reliance on survey-based rounds and enhance the timeliness and sensitivity of predictions. Another important direction is the extension of the framework to cross-country datasets. Financial distress manifests differently across macroeconomic regimes, regulatory environments, and cultural contexts. By testing models on data from multiple countries, it becomes possible to assess generalizability, adapt feature engineering strategies to new settings, and benchmark model stability across diverse economic structures. Such cross-country validation would also support the development of global early warning standards, allowing international agencies to scale solutions more effectively. Finally, future research should explore graph-based representations of household interdependencies. Financial distress is rarely an isolated phenomenon; households are embedded in networks of kinship, lending groups, supply chains, and digital finance ecosystems. Graph neural networks and related methods can capture spillover effects, such as how the distress of one household influences neighboring households or small enterprises. Incorporating such interdependencies would provide a more holistic view of vulnerability propagation, moving beyond individual-level prediction to system-level resilience modeling

## Conclusion

This study demonstrates that financial distress can be detected using sparse but timely socioeconomic and digital economy signals. By constructing a machine learning pipeline that incorporates data preprocessing, engineered temporal and interaction features, and ensemble-based classification models, the framework establishes a reliable foundation for early warning applications. The findings highlight that while binary classification of distress remains a low-signal challenge, multi-class severity prediction shows strong predictive power, especially when boosted ensemble models such as LightGBM and XGBoost are applied. A central contribution of the work lies in its emphasis on interpretability. The integration of SHAP-based explanations ensures that model outputs are not opaque, providing both global rankings of feature importance and local case-level narratives. These interpretability tools create a bridge between technical predictions and practical decision-making, fostering trust among policymakers, regulators, and households. From a deployment perspective, the framework charts a path toward policy-relevant applications. It illustrates how predictive analytics can shift intervention from reactive to proactive, allowing for earlier targeting of financial relief, regulatory monitoring, and household stabilization. Moreover, the scalability of the system ensures that, with richer datasets and integration of real-time feeds, it can operate as a continuous monitoring tool adaptable to both national and regional contexts. In sum, the research contributes to the growing body of evidence that machine learning can provide transparent, scalable, and operationally relevant early warning systems for financial distress. By demonstrating feasibility with limited but structured data, the study underscores the potential for future expansion into richer, higher-frequency, and network-aware models that can strengthen resilience in increasingly volatile financial environments.